\documentclass[arxiv]{melba}
\usepackage{mwe} % to get dummy images, only for the example
\usepackage{bbm}
\usepackage{amsmath,amsfonts}
\usepackage{booktabs}
\usepackage{multirow}
\usepackage{makecell}
\usepackage[table]{xcolor}
\usepackage{colortbl}
\usepackage{array}

\usepackage{hyperref}
\hypersetup{hidelinks,
	colorlinks=true,
	allcolors=black,
	pdfstartview=Fit,
	breaklinks=true}

\definecolor{baselinegray}{RGB}{238,238,238}
\definecolor{groupgray}{RGB}{242,242,242}
\definecolor{headerblue}{RGB}{221,229,239}
\definecolor{subheaderblue}{RGB}{242,246,250}

  \usepackage{xcolor}
  \definecolor{yesgreen}{RGB}{46,125,50}
  \definecolor{nored}{RGB}{198,40,40}
  \newcommand{\cmark}{\textcolor{yesgreen}{\checkmark}}
  \newcommand{\xmark}{\textcolor{nored}{$\times$}}

\melbaid{YYYY:NNN}  % This is provided upon by the publishing editor
\doi{10.59275/j.melba.2024-AAAA}

\melbaauthors{Yuan and Shujian}  % Note: this one is also used to set the pdf 'authors' metadata
\email{zuozhuliu@intl.zju.edu.cn}
\volume{3}
\firstpageno{1337}  % Communicated by the publishing editor
\melbayear{2026}  % The publication year
\datesubmitted{yyyy-m1-d1}  % Date submitted to MELBA: mm/yyyy
\datepublished{yyyy-m2-d2}  % Today's date: mm/yyyy

\melbaspecialissue{Medical Imaging with Deep Learning (MIDL) 2020}
\melbaspecialissueeditors{Marleen de Bruijne, Tal Arbel, Ismail Ben Ayed, Hervé Lombaert}

\ShortHeadings{MedStreamBench}{Yuan and Shujian}

\title{MedStreamBench: A Time-Aware Benchmark for Streaming and Proactive Medical Video Understanding}

\author{
	\firstname Yuan \surname Wang\aff{1,4},
	\name Shujian \surname Gao\aff{2,4},
    \name Songtao \surname Jiang\aff{1},
    \name Zhengyu \surname Hu\aff{3},
    \name Zuozhu \surname Liu\aff{1}
}
\affiliations{% <- trailing '%' to avoid unwanted indent
	\num 1 \addr College of Computer Science and Technology, Zhejiang University, Zhejiang, China \\
	\num 2 \addr Insititute of Trustworthy Embodied AI, Fudan University, Shanghai, China \\
	\num 3 \addr Department of Electrical and Computer Engineering, Carnegie Mellon University, Pennsylvania, USA \\
    \num 4 \addr Equal Contribution.
}

\abstract{
Existing medical video benchmarks primarily evaluate whether a model produces the correct answer, but rarely assess whether it answers at the \textbf{right time}. In real clinical settings, AI systems must decide not only what to predict, but also when to answer, defer judgment, or proactively raise alerts. This creates a critical gap between benchmark evaluation and deployment requirements. We present \textbf{MedStreamBench}, a benchmark for \textbf{time-aware medical video understanding}. MedStreamBench integrates \textbf{22 medical datasets} and \textbf{5,419 QA instances} across four temporal settings: retrospective, present, future, and proactive. Unlike conventional benchmarks that assume full-video access, MedStreamBench restricts models to temporally bounded evidence windows and supports both single-turn and streaming evaluation. We further introduce a \textbf{proactive monitoring} setting that requires models to determine whether and when clinically relevant alerts should be triggered. Beyond answer correctness, MedStreamBench evaluates temporal behavior through responsiveness and post-evidence stability. Experiments on leading general-purpose and medical vision-language models reveal a substantial gap between offline recognition and temporally grounded decision-making, with performance dropping markedly in streaming and proactive settings. Our benchmark is available at~\href{https://huggingface.co/datasets/Venn2024/MedStreamBench}{Venn2024/MedStreamBench}.}


\begin{document}

% top matter
\twocolumn[\maketitle]
% comment the preceedings and uncomment the following if the authors list + abstract is longer than one page
% \maketitle
% \twocolumn

   \begin{table*}[t]
      \centering
      \small
      \setlength{\tabcolsep}{5pt}
      \renewcommand{\arraystretch}{0.9}
      \caption{Comparison of MedStreamBench with representative surgical and medical video resources. We compare each resource by source
      breadth (\#DS), evaluation protocol, and coverage of the four temporal task modes.}
      \label{tab:benchmark_comparison}
      \begin{tabular}{p{2.9cm} r p{3.0cm} c c c c c c}
      \toprule
      \multirow{2}{*}{\textbf{Benchmark}} & \multirow{2}{*}{\textbf{\#DS}} & \multicolumn{3}{c}{\textbf{Protocol}} & \multicolumn{4}{c}
      {\textbf{Temporal Tasks}} \\
      \cmidrule(lr){3-5} \cmidrule(lr){6-9}
       &  & \textbf{Eval Unit} & \textbf{Prefix-Only} & \textbf{Streaming} & \textbf{Retrospective} & \textbf{Present} & \textbf{Future} &
       \textbf{Proactive} \\
      \midrule
      \rowcolor{groupgray}
      \multicolumn{9}{l}{\textbf{Classical Surgical Datasets}} \\
      Cholec80         & 1  & frame / video   & \xmark & \xmark & \cmark & \cmark & \xmark & \xmark \\
      M2CAI16-Workflow & 1  & frame / video   & \xmark & \xmark & \cmark & \cmark & \xmark & \xmark \\
      JIGSAWS          & 1  & gesture / video & \xmark & \xmark & \cmark & \cmark & \xmark & \xmark \\
      \midrule
      \rowcolor{groupgray}
      \multicolumn{9}{l}{\textbf{Recent Surgical / Medical Video Resources}} \\
      LLaVA-Surg       & 1  & clip / video    & \xmark & \xmark & \cmark & \cmark & \xmark & \xmark \\
      SurgPub-Video    & 1  & clip / video    & \xmark & \xmark & \cmark & \cmark & \xmark & \xmark \\
      SurgOnAir        & 1  & streaming video & \cmark & \cmark & \xmark & \cmark & \xmark & \xmark \\
      MedHorizon       & 1  & long video      & \xmark & \xmark & \cmark & \cmark & \xmark & \xmark \\
      \midrule
      \rowcolor{headerblue}
      \textbf{MedStreamBench (Ours)} & \textbf{22} & \textbf{long video / stream} & \textbf{\cmark} & \textbf{\cmark} & \textbf{\cmark} &
      \textbf{\cmark} & \textbf{\cmark} & \textbf{\cmark} \\
      \bottomrule
      \end{tabular}
  \end{table*}

% Introduction (or first section)
% \rule{\textwidth}{1pt}
\section{Background}
	\enluminure{M}{edical} video understanding is becoming increasingly important for surgical intelligence, endoscopic screening, procedural education, and real-time clinical decision support. Unlike static medical images, medical videos encode evolving procedural context: instruments enter and leave the field, anatomical exposure changes over time, lesions may appear only transiently, and safety-relevant events often depend on the temporal order of visual evidence. Consequently, a clinically useful medical vision--language model should not only recognize what is visible, but also determine \emph{when} sufficient evidence becomes available, \emph{when} to abstain, and \emph{when} to alert. This temporal discipline is essential in scenarios such as polyp screening, surgical workflow monitoring, view-quality assessment, and skill coaching, where premature conclusions can be misleading and delayed responses can reduce clinical utility.

	Recent video-language benchmarks have substantially advanced the evaluation of multimodal large language models on video understanding, including long-video reasoning and online video analysis~\citep{fu2024videomme,wu2024longvideobench,li2025ovobench,streamingbench2025}. However, most existing benchmarks assume either access to a complete video or a fixed clip, and primarily measure whether the final answer is semantically correct. Such offline evaluation is insufficient for medical streaming video analysis, where models must respect the information actually available at a specified time point. In parallel, medical and surgical datasets have enabled progress in phase recognition, instrument detection, action triplet recognition, polyp detection, capsule-endoscopy abnormality recognition, and surgical visual question answering~\citep{m2cai2016workflow,nwoye2022cholectriplet,smedsrud2021kvasircapsule,surgicalvqa2022,pitvqa2024,jiang2025hulu,wang2025v2t,wang2026beyond,jiang2025omniv}. Yet these resources typically evaluate recognition or question answering in an offline setting, and do not explicitly test whether a model can answer, defer, or alert at the correct moment under a strict evidence window.

	We introduce \textbf{MedStreamBench}, a diverse temporal benchmark for medical video question answering. To the best of our knowledge, MedStreamBench is the first benchmark designed to evaluate \emph{streaming} medical video understanding, and the first to formulate a \emph{proactive} medical video question-answering paradigm. In the streaming paradigm, the same question is asked over progressively growing visual prefixes, and a model must output \texttt{unanswerable} until the evidence becomes sufficient. In the proactive paradigm, the model continuously monitors a medical video stream and must respond with one of \texttt{no\_alert}, \texttt{uncertain}, or \texttt{alert: <reason>}, triggering an alert only when sufficient visual evidence appears. These two paradigms directly reflect clinically meaningful requirements: a model should avoid leaking future information, avoid unsupported early alarms, detect time-sensitive findings or workflow events promptly, and maintain stable behavior once evidence is available.

	As summarized in Table~\ref{tab:benchmark_comparison}, recent medical video resources such as LLaVA-Surg~\citep{li2024llava}, SurgPub-Video~\citep{li2026surgpub}, SurgOnAir~\citep{he2026surgonair}, and MedHorizon~\citep{du2026medhorizon} extend evaluation beyond classical phase- or clip-level recognition; however, they still provide limited coverage of prefix-only answering, streaming evaluation, or future and proactive temporal reasoning. MedStreamBench is designed to be broad and comprehensive. It integrates 22 medical video and image-sequence datasets spanning laparoscopic surgery, open surgery, robotic skill assessment, hysterectomy, pituitary surgery, bariatric bypass surgery, colonoscopy, gastrointestinal endoscopy, and capsule endoscopy. The benchmark contains 5{,}419 question-answering records, including 4{,}663 single-turn items and 756 streaming items, which expand to 3{,}369 per-round streaming inference jobs. Its temporal taxonomy covers four complementary modes: \emph{retrospective} questions about past evidence, \emph{present} questions about the current visual state, \emph{future} questions that require prediction or deferred answerability from partial evidence, and \emph{proactive} monitoring questions that require timely alerts. The benchmark also contains both closed and open-ended questions, and combines original annotations, category labels, hybrid label-grounded generation, and conservative model-assisted mining to cover both high-confidence label-based tasks and visually grounded open-ended reasoning.

    A central design principle of MedStreamBench is that every question is tied to an explicit visual evidence window. For single-turn tasks, the input is restricted to $[t_q - w,\, t_q]$, where $t_q$ is the query time and $w$ is the context window. For streaming future and proactive tasks, each round is evaluated only on $[t_q,\, t_c]$, where $t_c$ is the current time for that round. This design prevents temporal leakage and converts medical video question answering from an offline recognition problem into a time-aware decision problem. For open-ended questions, MedStreamBench further uses fine-grained judge rubrics that separately assess semantic correctness, evidence grounding, and safety, improving fairness and reducing ambiguity compared with a single coarse match criterion. In particular, the benchmark asks: given only the frames available so far, should the model answer, abstain, remain uncertain, or raise an alert?

\begin{figure*}[t]
			\centering
			\includegraphics[width=0.95\textwidth]{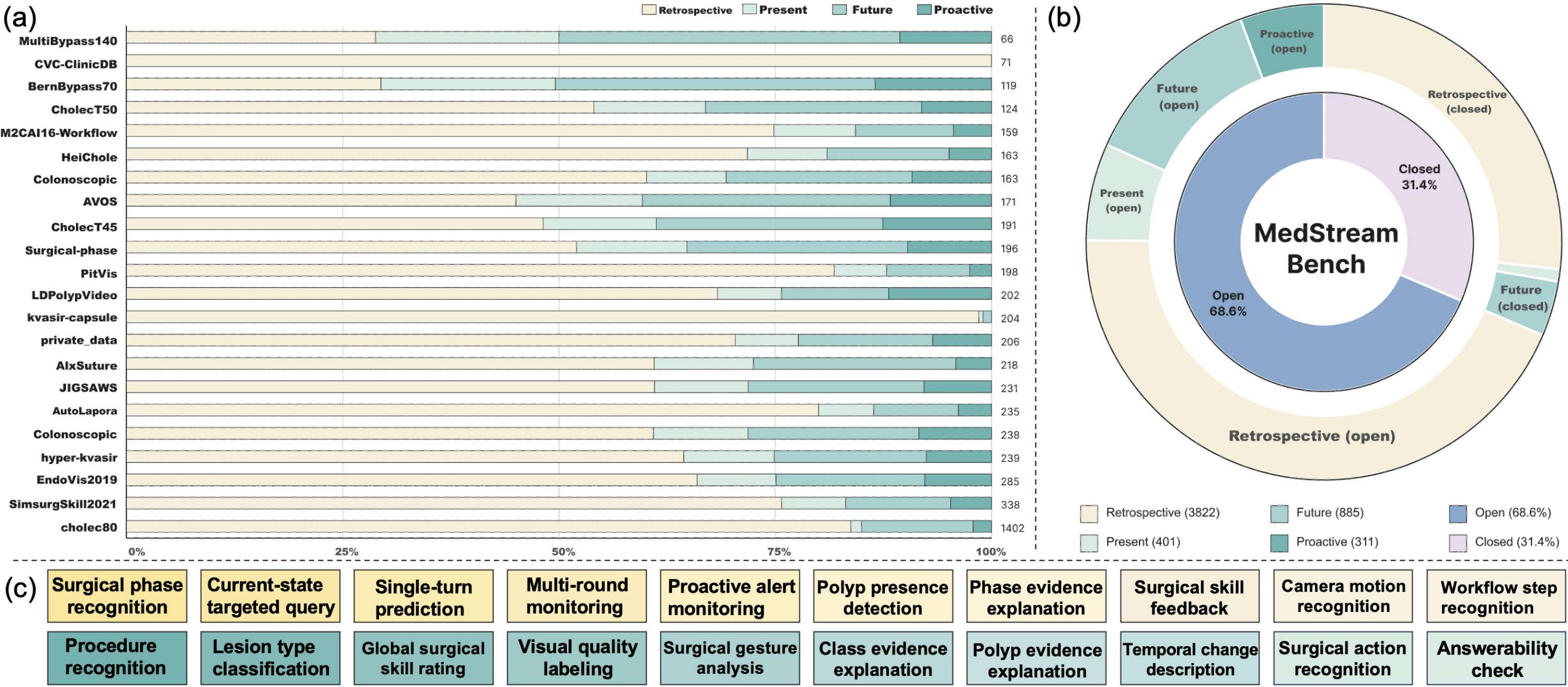}
			\caption{Overview of MedStreamBench as a resource. The benchmark integrates heterogeneous medical video and image-sequence datasets into a unified time-aware QA format with explicit evidence windows, multiple temporal modes, and both single-turn and streaming evaluation settings.}
			\label{fig:benchmark_overview}
		\end{figure*}

	To support scalable construction and reproducible evaluation, we develop two accompanying systems. First, the \textbf{MedStreamBench data engine} harmonizes heterogeneous medical datasets into a unified temporal QA schema. It converts heterogeneous medical data, including structured annotations, directory labels, and weakly labeled videos, into unified temporal QA items, and then validates and materializes the final dataset-scoped benchmark files. Second, the \textbf{MedStreamBench inference engine} expands benchmark records into per-window or per-round jobs, samples visual inputs under a fixed temporal protocol, supports local and API-based multimodal inference, and evaluates predictions using content correctness, streaming responsiveness, and post-evidence stability. This separation between data construction and inference makes the benchmark extensible to new datasets, models, and temporal protocols.

	Our contributions are threefold.
	\begin{itemize}
		\item \textbf{Time-aware formulation.} We propose a time-aware medical video QA formulation that unifies retrospective, present, future, streaming, and proactive evaluation under explicit evidence-window constraints.
		\item \textbf{Diverse benchmark.} We release a diverse benchmark built from 22 medical datasets, covering a wide range of procedures, anatomies, visual findings, task types, label sources, and temporal reasoning modes.
		\item \textbf{Reproducible engines.} We provide data and inference engines that enable reproducible benchmark generation, per-round streaming evaluation, and metrics for both answer correctness and temporal behavior.
	\end{itemize}
	% Together, MedStreamBench establishes a new evaluation setting for medical vision--language models: not only whether they can answer medical video questions, but whether they can answer or alert at the right time.

%%%%%%%%%%%%%%%%%%%%%%%%%%%%%%%%%%%%%%%%%%%%%%%%%%%%%%%%%%%%%%%%%%%%%%%%%%%
% Related works
%%%%%%%%%%%%%%%%%%%%%%%%%%%%%%%%%%%%%%%%%%%%%%%%%%%%%%%%%%%%%%%%%%%%%%%%%%%
% Make sure to put your work into context and include apporpriate citations.
% We do not have limits on citation counts.

%% 放任务类型，时序类型放到这

%% open-end rubrcis 单论多伦

\section{Summary}
\begin{figure*}[t]
    \centering
    \includegraphics[width=0.9\textwidth]{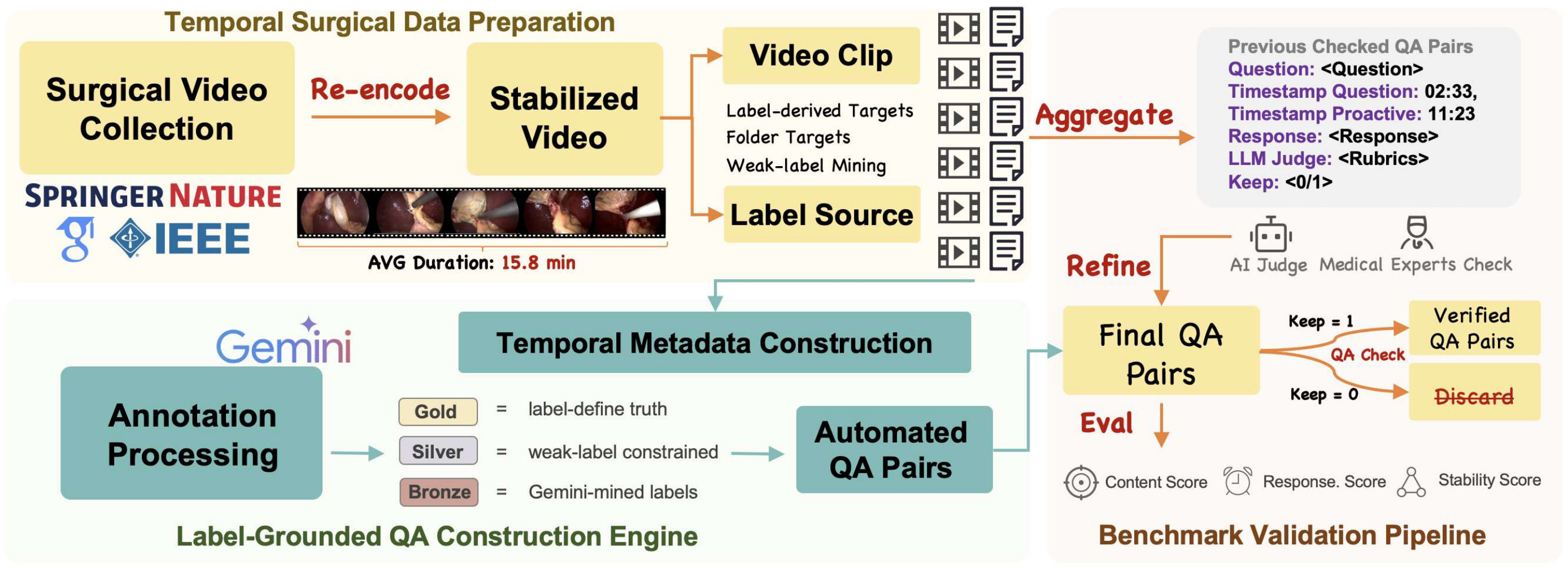}
    \caption{
    Overview of the MedStreamBench pipeline. The pipeline consists of three stages: temporal surgical data preparation, label-grounded QA construction, and benchmark validation. Source videos are standardized and paired with label-derived, folder-derived, or weakly mined supervision signals, which are converted into temporally aligned QA pairs with different quality levels. Candidate items are then aggregated, refined, and verified through AI-assisted judging and expert checking before final benchmark evaluation.
    }

    \label{fig:pipeline}
\end{figure*}

	MedStreamBench is a time-aware benchmark for medical video question answering with explicit evidence windows, four temporal modes, and both single-turn and streaming evaluation. As illustrated in Figure~\ref{fig:benchmark_overview} and Figure~\ref{fig:pipeline}, it integrates 22 source datasets into a unified JSONL schema that records the question, answer, temporal specification, task mode, response protocol, and label provenance. The current release contains 5{,}419 QA records, including 4{,}663 single-turn items and 756 streaming items, which expand to 3{,}369 round-level inference jobs during evaluation. The benchmark combines annotation-derived, directory-derived, hybrid, and conservatively model-assisted items to broaden coverage while retaining provenance information for downstream analysis.

	% \subsection{Dataset Coverage and Diversity}
	% 	MedStreamBench integrates heterogeneous medical visual sources spanning laparoscopic surgery, open surgery, robotic skill assessment, hysterectomy, pituitary surgery, bariatric bypass surgery, colonoscopy, gastrointestinal endoscopy, and capsule endoscopy. The source datasets vary substantially in annotation type and granularity, including phase labels, tool labels, action labels, segmentation masks, bounding boxes, skill metrics, category labels, and weakly supervised visual evidence. This breadth enables evaluation across multiple procedures, anatomies, visual targets, and temporal reasoning settings rather than restricting the benchmark to a single surgical workflow or diagnostic scenario.

	\subsection{AI-Readiness}
		The benchmark is organized in a unified machine-readable schema with explicit temporal metadata, including query time, input range, and round-level streaming structure where applicable. For open-ended questions, MedStreamBench additionally provides fine-grained judge rubrics that separately assess semantic correctness, evidence grounding, and safety, enabling more consistent LLM-as-judge evaluation than a single coarse matching criterion. In addition to benchmark files, MedStreamBench is paired with a standardized inference and evaluation workflow that fixes frame sampling, prompt construction, and temporal scoring, making it suitable for controlled comparison of model behavior under evidence-constrained medical video understanding.

\section{Discussion}
	\subsection{Strengths}
		MedStreamBench introduces a temporal evaluation setting that is largely absent from existing medical video resources. Its main strengths are explicit evidence-window control, broad multi-source coverage, and evaluation of both semantic correctness and temporally appropriate response behavior.

	\subsection{Limitations}
		The underlying source datasets differ substantially in annotation granularity, label quality, visual scope, and temporal density, and some benchmark items are derived from directory labels, hybrid generation, or weak supervision rather than fully expert-authored QA annotation. In addition, not all weakly supervised or model-assisted samples received exhaustive manual visual review, and the fixed frame-sampling protocol may underrepresent brief events.

	\subsection{Bias and Representativeness}
		As a multi-source benchmark, MedStreamBench inherits distributional biases from its constituent datasets, including differences in procedure types, devices, recording styles, annotation practices, and case distributions. Aggregate scores should therefore be interpreted together with dataset-level or domain-level analyses, especially when comparing behavior across surgical, endoscopic, and skill-assessment settings.

 \begin{table*}[t]
      \centering
      \scriptsize
      \setlength{\tabcolsep}{4pt}
      \renewcommand{\arraystretch}{1.08}
      \caption{Upstream access or license terms for the source datasets covered by MedStreamBench. The MedStreamBench release is intended
      to contain derived JSONL annotations and metadata only; original images and videos are obtained separately from the upstream sources.
      When a precise formal license could not be verified reliably from the source paper or project materials, we conservatively report
      `upstream terms' or `challenge terms' rather than infer a specific license.}
      \label{tab:source_access}
      \begin{tabular}{p{3.0cm} p{2.5cm} | p{2.8cm} p{2.5cm}| p{2.5cm} p{2.5cm}}
          \toprule
          \rowcolor{headerblue}
          \textbf{Dataset} & \textbf{Access} & \textbf{Dataset} & \textbf{Access} & \textbf{Dataset} & \textbf{Access} \\
          \midrule
          cholec80 & upstream terms & M2CAI16-Workflow & challenge terms & EndoVis2019 & challenge terms \\
          AutoLapora & public release & LDPolypVideo & upstream terms & PitVis & public release \\
          simsurgskill\_2021 & public release & kvasir-capsule & public release & hyper-kvasir & public release \\
          Colonoscopic-web & upstream terms & CVC-ClinicDB & public release & AlxSuture & upstream terms \\
          AVOS & public release & Surgical-phase & will release & JIGSAWS & public release \\
          CholecT45 & public release & CholecT50 & public release & HeiChole & challenge terms \\
          BernBypass70 & upstream terms & MultiBypass140 & upstream terms & private\_data & will release \\
          Colonoscopic-addi & upstream terms &  &  &  &  \\
          \bottomrule
      \end{tabular}
  \end{table*}

	\subsection{Responsible Use}
		MedStreamBench is intended as a research benchmark for evaluating time-aware medical video understanding systems. It should not be interpreted as a clinical decision-support system or as a substitute for expert judgment. Weakly supervised and model-assisted items are included to broaden evaluation coverage, but they should not be treated as expert-certified clinical labels. Any deployment-oriented use would require additional task-specific validation, prospective testing, and appropriate human oversight.

\section{Resource Availability}
	\subsection{Data Release}
		The benchmark is released as dataset-scoped JSONL files following a unified temporal QA schema and is hosted at~\href{https://huggingface.co/datasets/Venn2024/MedStreamBench}{Venn2024/MedStreamBench}. The release provides derived QA annotations and metadata under the~\href{https://choosealicense.com/licenses/apache-2.0/}{Apache License 2.0}, while the underlying image and video assets remain governed by the corresponding upstream sources.

		\subsection{Code Release and Access Policy}
		MedStreamBench is accompanied by a data construction pipeline and an inference/evaluation pipeline supporting benchmark generation, schema validation, frame sampling, prompt construction, streaming round expansion, and metric computation. Access to original videos or images remains subject to the terms, licenses, and availability conditions of the corresponding upstream datasets. The benchmark release therefore distributes derived QA JSONL files and associated metadata, but does not redistribute the original media unless permitted by the upstream source. When explicitly allowed by the corresponding upstream licenses or access terms, selected source media may also be redistributed or linked as part of the benchmark release, subject to those upstream conditions. Table~\ref{tab:source_access} summarizes the current access status and license terms for the 22 source datasets covered by MedStreamBench.

%%%%%%%%%%%%%%%

	\subsection{Intended Use}
		The intended use of MedStreamBench is the evaluation of medical vision--language models under temporally restricted visual evidence. Representative use cases include benchmark comparison, study of answerability under partial observation, evaluation of streaming response behavior, and research on proactive monitoring in medical video analysis. The resource is not intended to serve as a stand-alone clinical system.

\begin{figure*}[t]
    \centering
    \includegraphics[width=\textwidth]{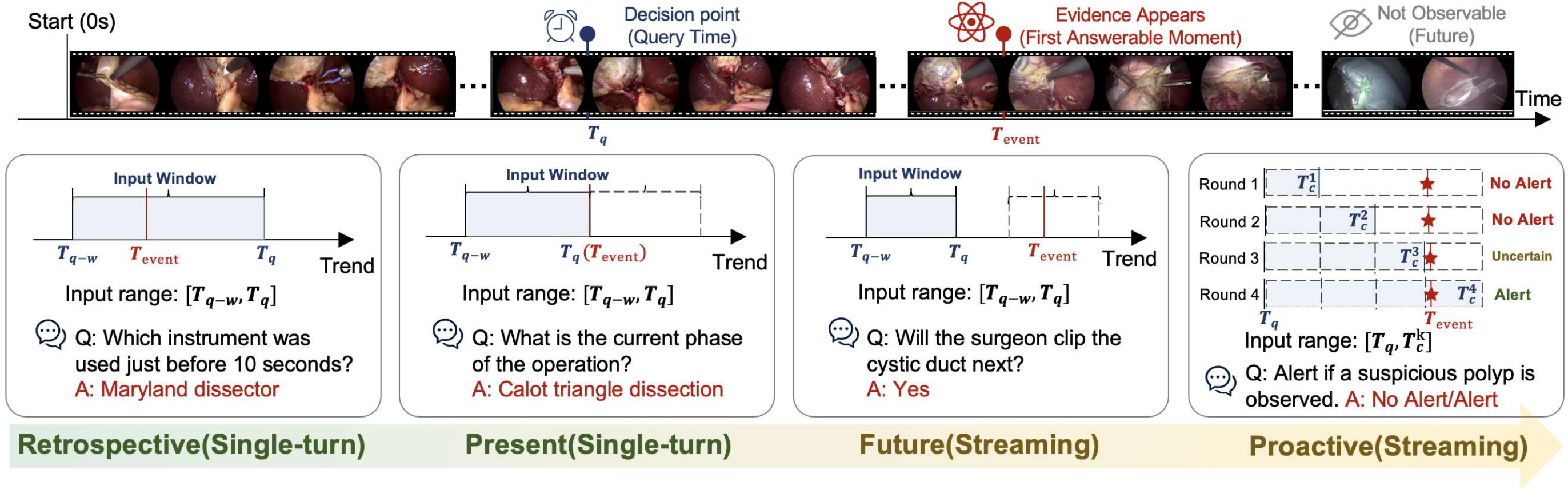}
    \caption{
    Illustration of the four temporal reasoning settings in MedStreamBench.
    Given a surgical video timeline, a decision point $T_q$ defines the latest observable evidence available to the model.
    Retrospective questions require understanding past events within the observation window $[T_{q-w},T_q]$, while present questions focus on the current surgical state near $T_q$.
    Future questions evaluate whether models can predict upcoming events before the first answerable moment $T_{\mathrm{event}}$ using only historical observations.
    Proactive streaming tasks further require the model to continuously monitor an evolving video prefix $[T_q,T_c^k]$ and trigger alerts only when sufficient evidence becomes visible.
    }
    \label{fig:four-mode}
\end{figure*}

% A methodological, model, or similar section often comes here.
\section{Methods}

	\subsection{Benchmark Overview}
		MedStreamBench is a time-aware medical video question-answering benchmark in which each item is associated with a visual source, an explicit temporal evidence window, a question, a reference answer, and metadata describing task type and label provenance. The release is stored as dataset-scoped JSONL files and covers both closed-choice and open-ended questions across retrospective, present, future, and proactive settings.

		Each item follows a compact unified schema:
		\begin{equation}
			x = (v, q, a, \tau, m, r, s),
		\end{equation}
		where $v$ denotes the visual source, $q$ is the question, $a$ is the answer or reference answer, $\tau$ denotes the temporal specification, $m$ denotes the task mode, $r$ contains evaluation rubrics or response protocols, and $s$ stores source and quality metadata. The temporal specification includes the query time $t_q$, the context window length $w$, and the input range $[t_s, t_e]$. For streaming tasks, each item additionally contains a sequence of rounds with current time $t_c$, round-specific input range, expected output, and answerability flag.

	\subsection{Data Construction Pipeline}
		The MedStreamBench data engine converts heterogeneous medical datasets into a unified temporal QA format. As illustrated in Figure~\ref{fig:pipeline}, the pipeline is organized into three aligned stages: temporal surgical data preparation, label-grounded QA construction, and benchmark validation.

		\paragraph{Temporal surgical data preparation.}
			We first standardize the source videos or image sequences and align them with available supervision signals. For structured datasets, these signals come from original annotations such as phases, tool labels, actions, masks, bounding boxes, or skill metrics; for other sources, they may also come from directory-defined categories or weak-label mining. Each target contains a visual source, a task specification, an answer signal, and an initial temporal window. For video datasets, timestamps are converted into seconds using dataset-specific metadata, whereas image-only datasets use the temporal range $[0, 0]$.

		\paragraph{Label-grounded QA construction.}
			The aligned supervision signals are then converted into temporally grounded QA items. For label-grounded tasks, deterministic targets are rewritten as natural-language closed or open-ended questions while preserving the answer. For weakly labeled or unlabeled videos, the engine conservatively mines candidate QA items from visible findings, procedural states, tool interactions, or screening alerts. To broaden temporal coverage, this stage also produces present, future, streaming, and proactive items rather than only retrospective recognition examples.

		\paragraph{Benchmark validation and materialization.}
			The resulting candidate items are aggregated, refined, and validated before release. The validator checks schema completeness, question and answer fields, formatting, media paths, temporal fields, and streaming rounds, and it enforces the temporal contract that single-turn items end at $t_q$ while streaming items use round-specific windows of the form $[t_q, t_c]$. Proactive items are additionally constrained to the response space \texttt{no\_alert}, \texttt{uncertain}, or \texttt{alert: <reason>}. Accepted items are then materialized into final dataset-scoped benchmark files for downstream inference and evaluation.

    		\begin{table}[t]
			\centering
			\footnotesize
			\setlength{\tabcolsep}{4pt}
			\renewcommand{\arraystretch}{1.10}
			\caption{Evaluation metrics used in MedStreamBench.}
			\label{tab:metric_definitions}
			\begin{tabular}{>{\raggedright\arraybackslash}p{0.22\columnwidth} >{\raggedright\arraybackslash}p{0.68\columnwidth}}
				\toprule
				\rowcolor{headerblue}
				\textbf{Metric} & \textbf{Interpretation} \\
				\midrule
				Content correctness ($C$) & Semantic correctness under the evidence-constrained input. For closed questions, it is based on normalized exact match with optional semantic fallback; for open questions, it combines correctness, evidence grounding, and safety. \\
				Responsiveness ($R$) & Temporal accuracy of the first positive answer or alert relative to the expected answerable or critical time. Early, delayed, and missed responses are penalized. \\
				Stability ($S$) & Post-evidence consistency after the item becomes answerable, measuring whether the model remains positive in later rounds rather than reverting to abstention or omission. \\
				Overall score ($O$) & Task-level summary score. For single-turn items, $O=C$; for streaming future and proactive items, $O=0.7\,C+0.3\,R$, prioritizing correctness while rewarding timely response. \\
				\bottomrule
			\end{tabular}
		\end{table}

	\subsection{Temporal Task Formulation}
		MedStreamBench defines four temporal modes and two task modes. Single-turn items produce one model call per item, whereas streaming items produce one model call per round.

		\paragraph{Retrospective and present single-turn QA.}
			For retrospective and present questions, the model receives a context window ending at the query time:
			\begin{equation}
				I = [\max(0, t_q - w),\ t_q].
			\end{equation}
			Retrospective questions ask about events or states within the window, whereas present questions focus on the visual state near $t_q$. If the available evidence is insufficient, the expected behavior is \texttt{unanswerable}.

		\paragraph{Future single-turn QA.}
			Future single-turn questions use the same input window:
			\begin{equation}
				I = [\max(0, t_q - w),\ t_q],
			\end{equation}
			but ask about an event or state after $t_q$. The model must reason from the available context only and return \texttt{unanswerable} when the future event is not yet predictable.

		\paragraph{Future streaming QA.}
			For future streaming questions, the item is evaluated over a sequence of rounds. At round $k$, the model receives only the prefix:
			\begin{equation}
				I_k = [t_q,\ t_c^{(k)}],
			\end{equation}
			where $t_c^{(k)}$ is the current time for that round. Before the evidence becomes sufficient, the expected output is \texttt{unanswerable}; once the target event becomes answerable, the model should answer and remain stable in subsequent rounds.

		\paragraph{Proactive streaming QA.}
			Proactive questions also use round-specific prefixes $[t_q, t_c^{(k)}]$, but the response space is constrained to:
			\begin{equation}
				y_k \in \{\texttt{no\_alert},\ \texttt{uncertain},\ \texttt{alert: <reason>}\}.
			\end{equation}
			The model should alert only when sufficient evidence appears, while using \texttt{no\_alert} or \texttt{uncertain} beforehand.

\begin{table*}[t]
\centering
\small
\setlength{\tabcolsep}{5.5pt}
\renewcommand{\arraystretch}{1.15}
\caption{Main benchmark results on MedStreamBench across the four temporal modes.
For retrospective and present settings, we report content accuracy only. For future and proactive settings, we report responsiveness and stability.}
% The overall score reports the benchmark-wide average item score using content for retrospective and present items and the task-specific score $0.7\,C+0.3\,R$ for future and proactive items. Higher is better for all metrics.}
\label{tab:main_temporal_results}
\resizebox{0.95\textwidth}{!}{%
\begin{tabular}{
l
@{\hspace{6pt}}
c
@{\hspace{10pt}}
c
@{\hspace{10pt}}
ccc
@{\hspace{10pt}}
ccc
@{\hspace{10pt}}
c
}

\toprule

\rowcolor{white}

\multirow{2}{*}{\textbf{Baseline}}

& \cellcolor{headerblue}\textbf{Retrospective}

& \cellcolor{headerblue}\textbf{Present}

& \multicolumn{3}{c}{\cellcolor{headerblue}\textbf{Future}}

& \multicolumn{3}{c}{\cellcolor{headerblue}\textbf{Proactive}}

& \cellcolor{headerblue}\textbf{Overall}

\\

\cmidrule(lr{8pt}){2-2}
\cmidrule(lr{8pt}){3-3}
\cmidrule(lr{8pt}){4-6}
\cmidrule(lr{8pt}){7-9}
\cmidrule(lr{8pt}){10-10}

& \cellcolor{subheaderblue}\textbf{Content}
& \cellcolor{subheaderblue}\textbf{Content}
& \cellcolor{subheaderblue}\textbf{Content}
& \cellcolor{subheaderblue}\textbf{Resp.}
& \cellcolor{subheaderblue}\textbf{Stab.}
& \cellcolor{subheaderblue}\textbf{Content}
& \cellcolor{subheaderblue}\textbf{Resp.}
& \cellcolor{subheaderblue}\textbf{Stab.}
& \cellcolor{subheaderblue}\textbf{Score}

\\

\midrule

\rowcolor{groupgray}
\multicolumn{10}{l}{\textbf{Closed-source}} \\
Gemini-2.5-Pro    & \textbf{0.3428} & \textbf{0.4140} & \textbf{0.4384} & \textbf{0.3636} & 0.6927 & 0.6158 & \textbf{0.4494} & \textbf{0.4742} & \textbf{0.3728} \\
% GPT-4.1           & -- & -- & -- & -- & -- & -- & -- & -- \\
% GPT-4o            & -- & -- & -- & -- & -- & -- & -- & -- \\
% Claude-3.7-Sonnet & -- & -- & -- & -- & -- & -- & -- & -- \\

\midrule
\rowcolor{groupgray}
\multicolumn{10}{l}{\textbf{Open-source}} \\
Qwen3-VL-4B        & 0.2255 & 0.3491 & 0.4046 & 0.1870 & 0.3415 & \textbf{0.6428} & 0.2301 & 0.2114 & 0.2701 \\
Qwen3-VL-8B        & 0.2008 & 0.2898 & 0.3260 & 0.0912 & 0.3293 & 0.6229 & 0.1499 & 0.1337 & 0.2324 \\
Qwen2.5-VL-7B      & 0.2413 & 0.3446 & 0.3656 & 0.0533 & 0.4635 & 0.6255 & 0.0547 & 0.0525 & 0.2662 \\
InternVL3.5-8B       & 0.2536 & 0.3461 & 0.3847 & 0.0710 & 0.3934 & 0.4441 & 0.0899 & 0.0813 & \textbf{0.2713} \\
LLaVA-OneVision-7B & 0.2498 & 0.2888 & 0.2890 & 0.0746 & 0.8142 & 0.5470 & 0.0579 & 0.0595 & 0.2572 \\

\midrule
\rowcolor{groupgray}
\multicolumn{10}{l}{\textbf{Medical-domain}} \\
HuluMed-4B & 0.2231 & 0.2868 & 0.4327 & 0.1456 & 0.2399 & 0.4742 & 0.2088 & 0.2248 & 0.2578 \\
HuluMed-7B & 0.2298 & 0.2798 & 0.3851 & 0.1425 & 0.1961 & 0.6258 & 0.0677 & 0.0723 & \textbf{0.2601} \\
Lingshu-7B & 0.2329 & 0.3302 & 0.3417 & 0.0953 & \textbf{0.7180} & 0.6302 & 0.0514 & 0.0514 & 0.2586 \\

\bottomrule
\end{tabular}
}

\vspace{2pt}
\footnotesize
\textbf{Content}: content accuracy.
\textbf{Resp.}: responsiveness score.
\textbf{Stab.}: stability score.
% \textbf{Overall}: benchmark-wide average item score with mode-specific task scoring.
\end{table*}

	\subsection{Inference Protocol}
		The MedStreamBench inference engine expands each benchmark item into one or more inference jobs. Single-turn items yield one job, whereas future and proactive streaming items yield one job per round.

		For video inputs, the engine samples frames uniformly at one frame every 2 seconds within the job-specific input range, including the endpoint when needed so that evidence at $t_q$ or $t_c$ remains visible. For image-sequence datasets, the input range is interpreted analogously as a frame-index interval. The textual prompt specifies the task mode, temporal mode, query time, input range, and the constraint that no evidence beyond the designated endpoint may be used. Future streaming prompts require \texttt{unanswerable} before sufficient evidence appears, while proactive prompts restrict the response space to \texttt{no\_alert}, \texttt{uncertain}, or \texttt{alert: <reason>}.

	\subsection{Evaluation Metrics}
		MedStreamBench evaluates both answer correctness and temporal behavior. Table~\ref{tab:metric_definitions} summarizes the interpretation of the reported metrics; below we specify the corresponding scoring rules.

		\paragraph{Content correctness.}
			For closed questions, predictions are evaluated by normalized exact match, with optional judge fallback for semantically equivalent mismatches. For open-ended questions, a rubric-based judge scores correctness, evidence grounding, and safety. The resulting score is normalized to $[0, 1]$:
			\begin{equation}
				C = \frac{s_{\mathrm{correct}} + s_{\mathrm{ground}} + s_{\mathrm{safety}}}{5},
			\end{equation}
			where $s_{\mathrm{correct}} \in \{0, 1, 2\}$, $s_{\mathrm{ground}} \in \{0, 1, 2\}$, and $s_{\mathrm{safety}} \in \{0, 1\}$. For streaming items, content correctness is averaged over rounds.

		\begin{figure*}[t]
			\centering
			\includegraphics[width=0.93\textwidth]{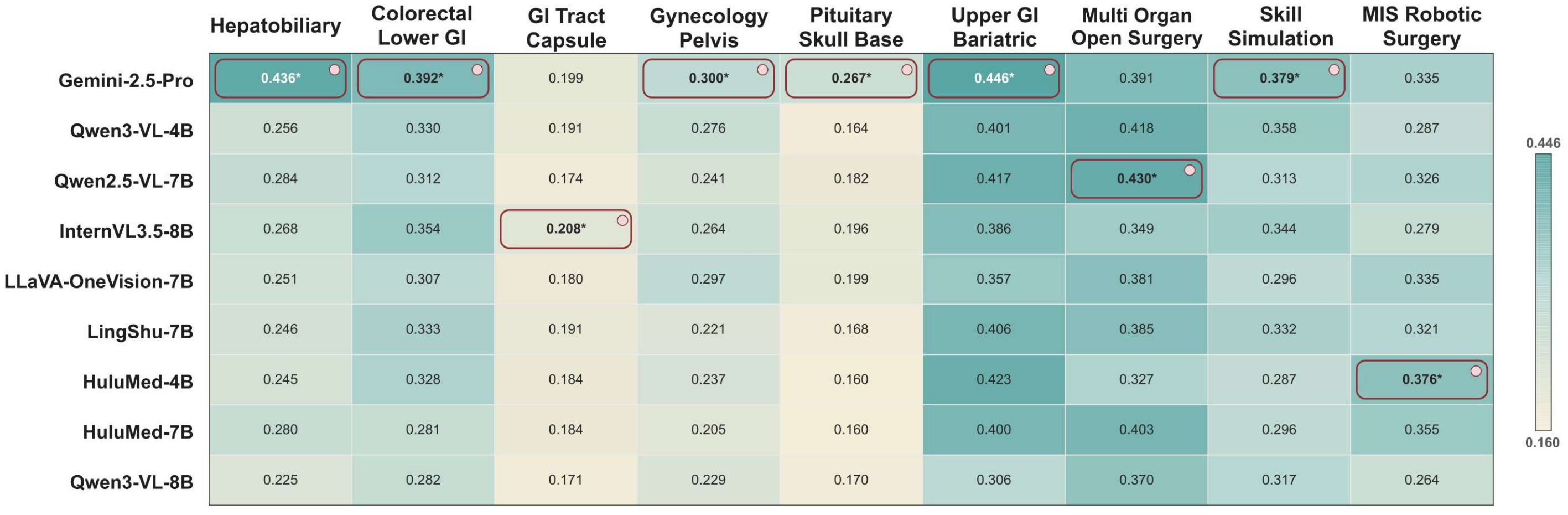}
			\caption{Organ-level aggregated content performance across baseline models. Each cell reports the sample-count-weighted average content score over datasets belonging to the same organ or procedural group. Darker teal indicates better performance. The best model in each column is marked with a star and highlighted by a thin outline.}
			\label{fig:organ_heatmap}
		\end{figure*}
		\paragraph{Responsiveness.}
			For future and proactive streaming items, responsiveness measures whether the first positive response occurs near the expected response time. For future streaming tasks, a positive response is any answer other than \texttt{unanswerable}; for proactive tasks, it is an \texttt{alert}. Let $t^*$ be the expected response time, $\hat{t}$ the first predicted positive time, and $\delta$ the tolerance. A prediction receives the maximum responsiveness score if
			\begin{equation}
				|\hat{t} - t^*| \leq \delta.
			\end{equation}
			Missed responses receive zero responsiveness, and premature or delayed responses are penalized according to their temporal distance from $t^*$.

		\paragraph{Stability.}
			For streaming tasks, stability measures whether the model remains positive after the item becomes answerable:
			\begin{equation}
				S = \frac{1}{|\mathcal{K}_{\geq t^*}|} \sum_{k \in \mathcal{K}_{\geq t^*}} \mathbbm{1}[y_k \text{ is positive}],
			\end{equation}
			where $\mathcal{K}_{\geq t^*}$ denotes rounds at or after the expected response time.

		\paragraph{Overall score.}
			For single-turn items, the overall score is the content score:
			\begin{equation}
				O = C.
			\end{equation}
			For streaming future and proactive items, the overall score combines correctness and responsiveness:
			\begin{equation}
				O = 0.7\,C + 0.3\,R,
			\end{equation}
			where $R$ is the responsiveness score.

		% \paragraph{Aggregation.}
		% 	Scores are aggregated overall and by dataset, temporal mode, question type, and task mode, allowing MedStreamBench to distinguish semantic errors from temporal failures under a unified evaluation protocol.

\section{Validation}
			\subsection{Baseline description}
			Table~\ref{tab:main_temporal_results} reports representative baselines from three groups: a closed-source model~\citep{team2023gemini}, five general-domain open-source models~\citep{bai2025qwen3,zhu2025internvl3}, and three medical-domain models~\citep{xu2025lingshu,jiang2025hulu}. All experiments were conducted on NVIDIA H100 GPUs under the same benchmark protocol.

		\subsection{Main Results}
		Table~\ref{tab:main_temporal_results} summarizes initial baseline results on MedStreamBench. Together with the metric definitions in Table~\ref{tab:metric_definitions}, it separates semantic accuracy from temporal responsiveness and stability across retrospective, present, future, and proactive settings. This separation makes it possible to distinguish models that perform well in offline recognition from models that are also temporally disciplined under prefix-only evidence.
			The overall score provides a compact benchmark-level summary by aggregating item scores under the task-specific evaluation rule defined in Section~4.2. Across the evaluated models, Gemini-2.5-Pro achieves the strongest overall performance, while InternVL3.5-8B is the best-performing open-source model and Lingshu-7B yields the strongest overall score among the medical-domain models. These aggregate results are useful for high-level comparison, but they should be interpreted together with the mode-specific columns because temporal strengths are not uniformly distributed across 4 task modes.

            \begin{figure}[t]
			\centering
			\includegraphics[width=0.79\columnwidth]{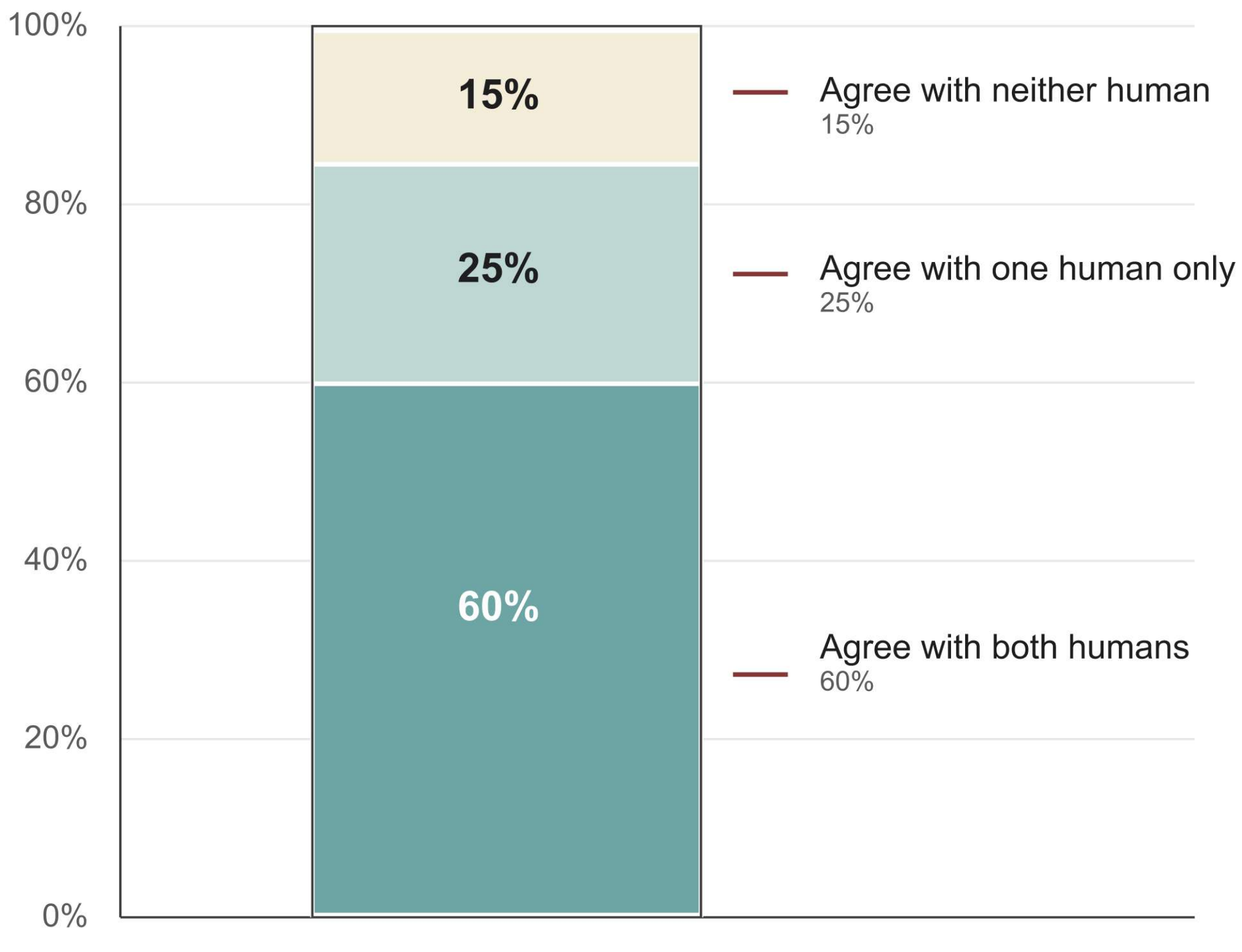}
			\caption{Manual spot-check agreement between the AI-generated answer and two human annotators.}
			\label{fig:human_check}
		\end{figure}

			\subsection{Organ-level analysis}
			Figure~\ref{fig:organ_heatmap} aggregates dataset-level content performance by organ or procedural domain using sample-count-weighted averaging. Gemini-2.5-Pro achieves the strongest performance across most organ groups, whereas the best open or medical-domain models emerge in more specialized regions, including InternVL3.5-8B on the GI tract and capsule group, Qwen2.5-VL-7B on multi-organ open surgery, and HuluMed-4B on MIS robotic surgery. This heterogeneous pattern shows that benchmark performance is not uniform across clinical domains and motivates organ-level analysis beyond a single global average. Figure~\ref{fig:benchmark_overview} and Figure~\ref{fig:four-mode} provide the corresponding resource-level and temporal-task context for interpreting these benchmark-level quantitative results.

\subsection{Human Check}
		As a preliminary quality check for model-assisted items, we compared AI-generated answers with judgments from two independent human annotators on a random sample. Figure~\ref{fig:human_check} shows agreement with both annotators in 60\% of cases, with one annotator in 25\%, and with neither annotator in 15\%.

%%%%%%%%%%%%%%%%%%%%%%%%%%%%%%%%%%%%%%%%%%%%%%%%%%%%%%%%%%%%%%%%%%%%%%%
% Mandatory Sections. Please complete, especially for final publication
%%%%%%%%%%%%%%%%%%%%%%%%%%%%%%%%%%%%%%%%%%%%%%%%%%%%%%%%%%%%%%%%%%%%%%%

% Acknowledgements.
% Please include any funding, intellectual contributions not included in the authorship, and any other acknowledgements.
\acks{This work is supported by the "Pioneer" and "Leading Goose" R\&D Program of Zhejiang (Grant no. 2025C01128), and the ZJU-Angelalign R\&D Center for Intelligence Healthcare.}

% Ethical Standards.
% Please edit with the appropriate ethics considerations for your work. Include any pertinent IRB information, etc.
%
% Please note that the submission requirements included:
% The work presented must follow appropriate ethical standards in conducting research and writing the manuscript, following all applicable laws and regulations regarding treatment of animals or human subjects.
% \ethics{MedStreamBench is intended for research use in benchmark construction and evaluation of time-aware medical video understanding systems. The resource aggregates derived annotations and metadata from public or otherwise permitted upstream sources, and access to original source media remains subject to the terms of the corresponding datasets. The benchmark should not be used as a substitute for clinical judgment or as a stand-alone clinical decision system. Any deployment-oriented use would require additional validation, domain-specific review, and appropriate human oversight.}

% Conflict of Interest
% Declaration of possible conflicts of interest: Authors must disclose any financial, organisational, commercial or personal conflicts of interest that might bias their work.
% If no conflicts, please say "We declare we don't have conflicts of interest."
\coi{We declare we don't have conflicts of interest.}

% Data availability
% \data{The current MedStreamBench benchmark release is available at \url{https://huggingface.co/datasets/Venn2024/MedStreamBench}. The release includes derived benchmark annotations and temporal metadata. Original videos and images are not necessarily redistributed and may need to be obtained separately from the corresponding upstream datasets under their respective access conditions and licenses.}

\end{document}